\documentclass[10pt,twocolumn,letterpaper]{article}

\usepackage{overpic}
\usepackage{float}
\usepackage{bm}
\usepackage{multicol}
\usepackage{multirow}
\usepackage{pifont}
\usepackage{color}
\usepackage{overpic}
\usepackage{arydshln}
\usepackage{makecell}
\usepackage{times}
\usepackage{epsfig}
\usepackage[T1]{fontenc}
\usepackage{hhline}
\usepackage{pifont}
\usepackage[shortlabels,inline]{enumitem}
\usepackage{colortbl}
\usepackage{verbatim}
\usepackage{bm}
\usepackage{fancyhdr}
\usepackage{multirow}
\usepackage[pagenumbers]{iccv} % To force page numbers, e.g. for an arXiv version

\definecolor{iccvblue}{rgb}{0.21,0.49,0.74}
\usepackage[pagebackref,breaklinks,colorlinks,allcolors=iccvblue]{hyperref}

\title{VideoGen-Eval: Agent-based System for Video Generation Evaluation}

\author{Yuhang Yang$^{1}$, Ke Fan$^{2}$, Shangkun Sun$^{3}$, Hongxiang Li$^{3}$, Ailing Zeng$^{4,\dagger}$, Feilin Han$^{5}$ \\ Wei Zhai$^{1,\dagger}$, Wei Liu$^{4}$, Yang Cao$^{1}$, Zheng-Jun Zha$^{1}$\\
{$^{1}$~USTC \textit{} $^{2}$~SJTU \textit{} $^{3}$~PKUSZ \textit{} $^{4}$~Tencent \textit{} $^{5}$~BFA}\\
$\dagger$Corresponding Author\\
\href{https://github.com/AILab-CVC/VideoGen-Eval}{https://github.com/AILab-CVC/VideoGen-Eval}
}

\begin{document}
\maketitle
\begin{abstract}
The rapid advancement of video generation has rendered existing evaluation systems inadequate for assessing state-of-the-art models, primarily due to simple prompts that cannot showcase the model's capabilities, fixed evaluation operators struggling with Out-of-Distribution (OOD) cases, and misalignment between computed metrics and human preferences. To bridge the gap, we propose VideoGen-Eval, an agent evaluation system that integrates LLM-based content structuring, MLLM-based content judgment, and patch tools designed for temporal-dense dimensions, to achieve a \textbf{dynamic}, \textbf{flexible}, and \textbf{expandable} video generation evaluation. Additionally, we introduce a video generation benchmark to evaluate existing cutting-edge models and verify the effectiveness of our evaluation system. It comprises 700 structured, content-rich prompts (both T2V and I2V) and over 12,000 videos generated by 20+ models, among them, 8 cutting-edge models are selected as quantitative evaluation for the agent and human. Extensive experiments validate that our proposed agent-based evaluation system demonstrates strong alignment with human preferences and reliably completes the evaluation, as well as the diversity and richness of the benchmark.
\end{abstract}

\section{Introduction}
\label{sec:intro}
High-quality video generation, including text-to-video (T2V) and image-to-video (I2V), plays a crucial role in content creation. With Sora \cite{sora} pioneering large-scale video generation, recent models \cite{hunyuan,seaweed,mochi,cogvideox,pika,wan2.1} have significantly advanced the field, achieving higher resolutions, more natural motion, and improved instruction alignment. However, video evaluation has failed to keep pace with these rapid advancements. Unlike images and text, videos exhibit greater spatiotemporal complexity, making evaluation significantly more challenging. The field urgently requires a flexible and extensible framework capable of assessing the evolving landscape of video generation.

\begin{figure}[t]
	\centering
        \scriptsize
	\begin{overpic}[width=1.\linewidth]{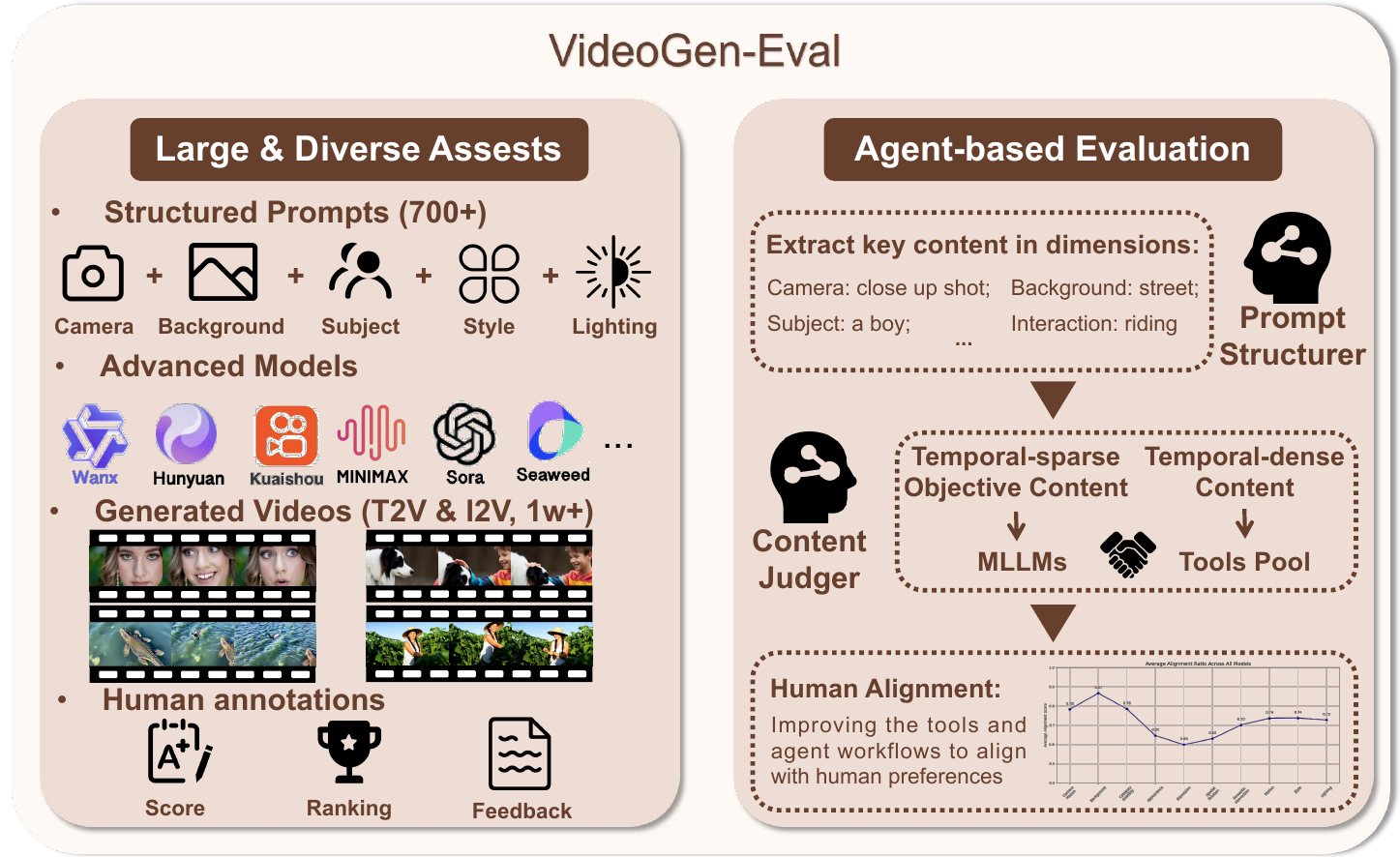}
 
	\end{overpic}
	\caption{\textbf{VideoGen-Eval.} Our benchmark includes structured prompts with rich content, large-scale results generated by multiple cutting-edge models, and human annotations. We also propose an agent-based dynamic evaluation system that can reliably complete the evaluation and adapt to human preferences.}
 \label{fig:teaser}
\end{figure}

\begin{figure}[t]
	\centering
        \scriptsize
	\begin{overpic}[width=1.\linewidth]{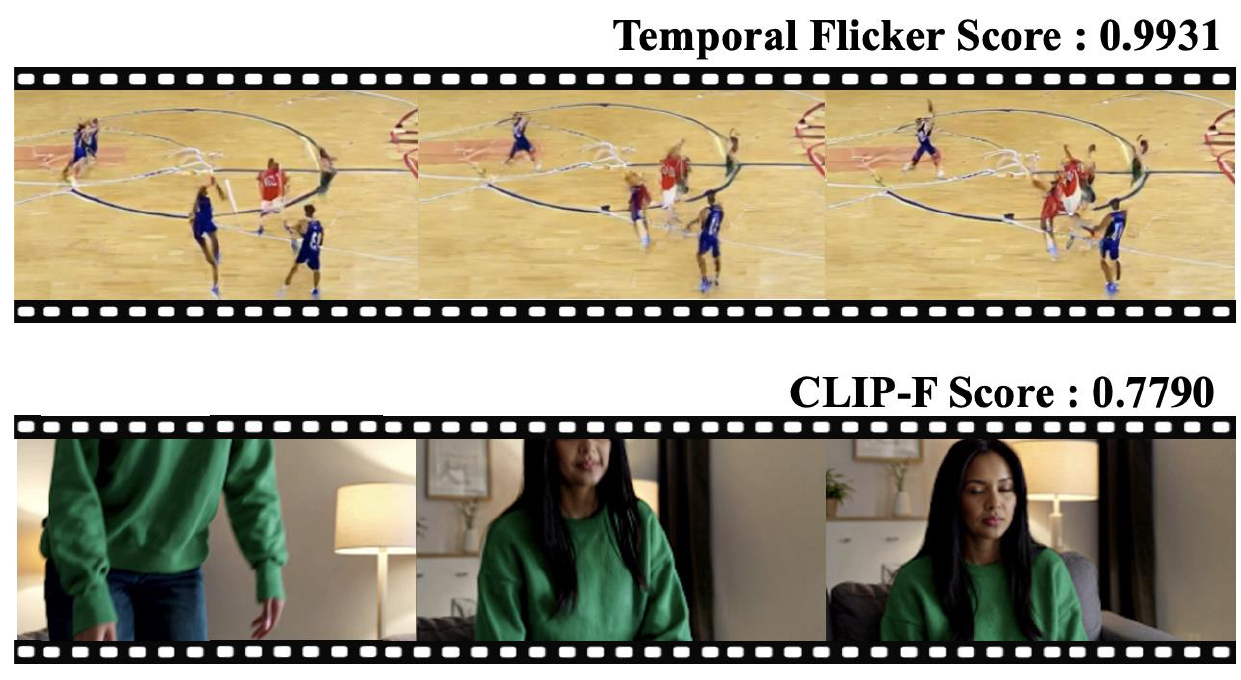}
 
	\end{overpic}
	\caption{The existing benchmarks adopt unreasonable evaluation operators, such as assigning high scores to videos with noticeable flickering while penalizing slight camera movements with disproportionately low semantic consistency scores.}
 \label{fig:bad_case}
\end{figure}

Some works \cite{vbench,t2vcompbench,vgeneval,wang2024aigvassessorbenchmarkingevaluatingperceptual,cao2024physgame,liu2025uve,evalcrafter,liu2023fetv,t2vqa} have explored the evaluation of video generation, but they typically suffer from the following limitations:
\textbf{1)} Non-structural prompts with simplistic content. Existing benchmarks rely on unstructured prompts that encapsulate only basic semantic concepts, \emph{e.g.}, “A panda standing on a surfboard in the ocean at sunset” \cite{vbench}, lacking explicit instructions for movement and key objective dimensions. Such simple prompts are insufficient to evaluate the capabilities of advanced video generation models~\cite{wan2.1,kling,seaweed,sora}. \textbf{2)} Fixed definitions of evaluation dimensions. Current benchmarks categorize evaluation dimensions based on video quality and content, yet these definitions are inherently open-ended and require different evaluation operators for distinct aspects. This complexity demands a more adaptable evaluation framework. \textbf{3)} Static and inflexible evaluation operators. Existing evaluation systems are predefined, fixed, and static, making them incapable of handling the varying complexity of generated videos. Operators trained on specific datasets struggle with out-of-distribution (OOD) content, leading to unreliable assessments. As shown in Fig. \ref{fig:bad_case}, the upper video exhibits strong flickering but receives a high score, while in the bottom row, slight camera motion causes large deviations in semantic score calculations, undermining the robustness of current evaluation methods.

Considering these limitations, we propose VideoGen-Eval, an agent-based system for dynamically evaluating the video generation, as illustrated in Fig. \ref{fig:teaser}. We generate over 12,000 generated video assets from over 20 commercial and open-source models \cite{gen3,kling,vidu,luma,hailuo,wan2.1,cogvideox,easyanimate,opensora,opensora-plan,mochi,seaweed,pika,pixeldance,hunyuan,sora,jin2024pyramidal} using the designed over 700 structured prompts with wide coverage of world concepts. To overcome the shortcomings of existing benchmarks and establish a reliable evaluation framework, we focus on three key aspects: 1) \emph{prompt design}; 2) \emph{evaluation protocols}; and 3) \emph{validation methodologies}.
Firstly, we construct structured prompts to ensure they encompass five essential components: camera, background, subject, style, and lighting. The subject dimension further integrates key attributes such as semantics, quantity, appearance, motion, spatial relations, etc. This structured manner enhances the richness of semantic concepts and motion descriptions in input prompts, allowing each test case to be assessed more content. This also better reflects the performance boundaries of current advanced models.

Secondly, we introduce a dynamic agent system for the evaluation of generated videos, comprising three key components: 1) a Large Language Model (LLM)-based content structurer \cite{gpt4,qwen}; 2) a Multimodal Large Language Model (MLLM)-based content judger \cite{Qwen-VL,gpt4}; and 3) a tools pool containing operators for assessing temporally dense dimensions. To ensure a clearer evaluation process, we leverage the strong comprehension and decomposition capabilities of LLMs to extract key content and concepts from input prompts, structuring them into distinct evaluation components. This approach ensures that evaluation criteria remain well-defined and adaptable, allowing the system to determine relevant evaluation aspects based on each input prompt. For the extracted evaluation content, the agent system employs MLLMs to verify whether the specified dimensional content has been correctly generated. Unlike previous methods that rely on MLLMs to directly assign scores \cite{t2vcompbench}, our approach shifts the role of MLLMs from subjective scoring to objective judgment, significantly enhancing evaluation reliability. Furthermore, to address MLLMs' limitations in evaluating temporal-dense dimensions \cite{cai2024temporalbench}, we introduce and improve some temporal-dense operators~\cite{amt,raft,dinov2} as tools. The agent system strategically invokes these operators to compute dimension-specific scores, thereby improving assessments of temporal attributes. This dynamic system, which adds patch operators on top of the fundamental understanding capabilities of MLLMs, enhances the reliability of evaluations.

Finally, to verify the alignment degree between the evaluation agent system and human preferences, we employ 20 annotators, including professional film practitioners, to annotate 5,500+ videos generated by 8 advanced models \cite{sora,kling,hunyuan,wan2.1,hailuo,runway2024gen3,seaweed}, including the judgment of whether the instructed content is generated and corresponding detailed feedback (potentially support post-training \cite{liu2025improving}). Note that we only select the latest 8 advanced models for final evaluation to ensure the advanced nature.

The contributions are summarized as follows:
\begin{itemize}[leftmargin=15pt,topsep=0pt,itemsep=2pt]
    \item[\textbf{1)}] We introduce a dynamic agent system for video generation evaluation, which orchestrates LLM-based content structuring, MLLM-based judging, and operator-based patch tools to achieve a dynamic, flexible, and expandable evaluation.
    \item[\textbf{2)}] We provide $12,000+$ videos generated by $20$+ open-source and commercial models with $700$ structured prompts, covering both T2V and I2V tasks. It provides numerous samples for observing the generated patterns of various models. Besides, 8 advanced models with human annotations are selected for quantitative tests, building the cutting-edge video generation benchmark.
    \item[\textbf{3)}] Experimental results demonstrate that our proposed agent-based system achieves better alignment with human preferences compared to existing static evaluation methods, leading to a more robust and reliable video generation evaluation.
\end{itemize}

\section{Related Work}
\label{sec:formatting}
\subsection{Methods for Video Generation}  
Controllable video generation has witnessed significant progress with the emergence of diffusion models. 
Since~\cite{make-a-video} extended the success of text-to-image to video generation, subsequent studies~\cite{lavie,modelscope,svd,videocrafter,opensora,opensora-plan,li2024dispose} have explored various UNet-based pipelines and incorporated more conditions such as image and video. 
Despite these efforts, these early models face persistent challenges: basic inconsistency \eg, flickering, along with limitations in resolution, aesthetic quality, and duration. Furthermore, their capabilities remained constrained to simple textual prompts and rudimentary motion patterns. 
A paradigm shift occurs with Sora~\cite{sora}, which demonstrates unprecedented capabilities through DiT-based architectures and sophisticated data curation strategies. 
This breakthrough catalyzes a wave of new-generation models~\cite{hunyuan,luma,hailuo,tongyi,pika,cogvideox} that substantially mitigate flickering issues while achieving higher resolution, cinema-grade aesthetics, and minute-long coherent narratives. These advanced models now support intricate textual instructions, multi-shot scene transitions, and fine-grained motion control—capabilities previously deemed challenging. 
Nevertheless, with the rapid advancement of video generation models, corresponding advanced benchmarks and measurements remain critically lacking, and the establishment of appropriate evaluation methodologies continues to present a significant challenge.

\subsection{Benchmarks for Video Generation}
Prior works have involved considerable exploration of benchmarks for evaluating generated videos~\cite{chivileva2023measuring, genaibench, chronobench, t2vcomb, t2vscore,vgeneval}.
EvalCrafter~\cite{evalcrafter} constructs the ECTV benchmark using 700 prompts and 8 models, evaluating the results with 18 objective metrics. 
FETV~\cite{liu2023fetv} utilizes 619 prompts and 4 T2V models to create a dataset of 2,476 videos. 
VBench~\cite{huang2023vbench} employs 1,746 prompts and 4 T2V models to generate 6,984 videos, which are evaluated using 16 objective metrics. 
T2VQA-DB~\cite{kou2024subjectivealigneddatasetmetrictexttovideo} built a quality assessment dataset for text-driven videos using 10 models and 1,000 prompts. However, current benchmarks still face the following challenges:  
(1) Lack of structured text prompts, leading to insufficient information density and difficulty in comprehensive or dimension-specific evaluation, as shown in Table~\ref{tab:prompt_comparison}; 
(2) Gap with current generative models. Early-generation models are widely used in those benchmarks, which face quite different challenges compared with concurrent models, as mentioned earlier; 
(3) Lack of multi-modal control evaluation. Among the benchmarks mentioned above, only VBench++ \cite{huang2024vbench++} includes additional evaluations for other control conditions such as image, while others mainly support evaluations for text-driven models.

\subsection{Measurements for Video Generation}
Existing measurements can be categorized into two groups: objective metrics and human-aligned subjective metrics. \textbf{Objective metrics} typically assess a specific dimension via pre-trained priors. For instance, some methods utilize averaged frame-wise similarity metrics to calculate temporal consistency~\cite{clip,lpips,dino,psnr,ssim,is}. Some leverage low-level motion analysis through optical flow~\cite{rave, flatten, skflow, streamflow}, frame interpolation~\cite{amt}, or tracking-based methods~\cite{t2vcompbench, cotracker} to measure the temporal variations. Some adopt vision-language models~\cite{umt, clip, li2022blip} to measure semantic alignment between generated content and multi-modal controls. 
Notwithstanding their merits, these measurements face challenges: 1) the \textit{smoothing effect} in frame-level computations tends to obscure spatial-temporal localized anomalies; 2) inherent bias introduced by single-dimensional metrics. For instance, CLIP-score favors static scenes over legitimate motion patterns; 3) lack of explicit alignment with human perceptions. 
\textbf{Human-aligned metrics} typically construct datasets with human feedback, and then train quality assessment models based on them, such as video assessment models~\cite{dover, trivqa, t2vqa, vebench, wu2023qalign} and some image-oriented assessment models~\cite{pickscore, hps, hpsv2, imagereward, iebench} that calculate the average of the frame-level subjective-aligned score. The Evaluation Agent \cite{zhang2024evaluation} is most similar to ours, which integrates the proposal and execution stages using an agent and completes the evaluation. In this work, we utilize cutting-edge powerful generative models to build a high-quality dataset with human annotations and establish a multi-modal agent for comprehensive evaluation.

\begin{table}[t]
\small
\centering
\scalebox{0.95}{
\begin{tabular}{lcc}
\toprule
Benchmark & \# Words & \# Characters \\ \midrule
GenAI-Bench~\cite{genaibench}	        & 7.25	& 39.30 \\
VBench~\cite{vbench}	& 7.64	& 41.95 \\
FETV~\cite{liu2023fetv}	& 10.94	& 59.60\\
T2V-CompBench~\cite{t2vcompbench}	& 10.42	& 56.42	\\
T2VScore~\cite{t2vscore} & 12.31 & 67.59 \\
EvalCrafter~\cite{evalcrafter}	&12.33	& 69.77 \\
T2VQA-DB~\cite{t2vqa}	        & 12.32	& 76.22	 \\
%ChronoMagic-Bench~\cite{chronobench} & 43.87 & 289.29 \\
\textbf{Ours}	& \textbf{57.29}	& \textbf{449.99} \\
\bottomrule
\end{tabular}
}
\caption{Comparison of average word and character counts in each prompt across different benchmarks.}
\label{tab:prompt_comparison}
\end{table}
\section{Evaluation Data Construction}
Our dataset comprises two components. The first part consists of large-scale generated assets: We construct 700 prompts (including 400 text-to-video and 300 image-to-video prompts) based on multiple dimensions and user inputs. These prompts are processed by over 20 models (both open-source and commercial), yielding over 12,000 generated videos. This corpus enables the analysis of generation patterns for specific models, supporting post-training \cite{liu2025improving} or enhancing perception models' understanding of synthetic contents. The second part mainly contains 8 advanced model results for quantitative testing, for this part, we formulate the following principles: \textbf{1)} Appropriate amount of test cases: as model scales grow exponentially, the computational overhead of evaluations also increases. The number of test cases should be controlled within a range; \textbf{2)} Rich content: while the quantity of test cases is limited, the prompt could be structured to involve more concepts, increasing the evaluation dimensions and richness; \textbf{3)} Cutting-edge models: given rapid advances in video generation, the benchmark should specifically evaluate advanced models to map their capability boundaries and identify performance ceilings of the field.

\begin{figure}[t]
	\centering
        \scriptsize
	\begin{overpic}[width=1.\linewidth]{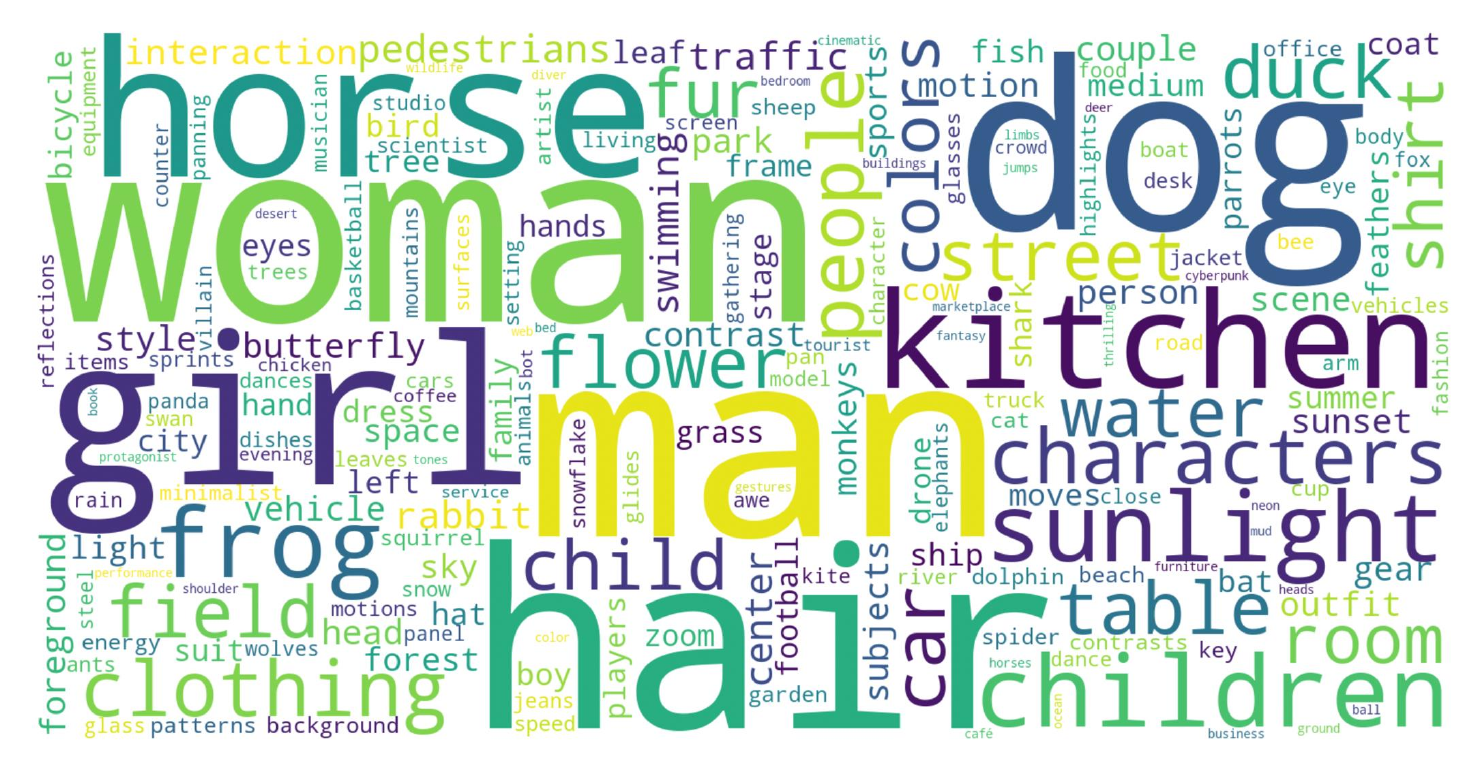}
 
	\end{overpic}
	\caption{Statistics of the word cloud in the collected prompts.}
 \label{fig:prompt_freq}
\end{figure}

Based on the above principles, we select a portion from the first part of 700 prompts and integrated prompts from multiple benchmarks (Vbench \cite{vbench}, MovieGen bench \cite{polyak2024moviegencastmedia}, T2V-CompBench \cite{t2vcompbench}) to construct 700 content-rich prompts. Each prompt contains five key components: camera movement, background description, subject description, style, and lighting description. Note that some benchmark prompts do not cover all these components, so we use GPT-4o \cite{gpt4} to structurally expand the raw prompts, ensuring that each one contains these components. The word frequency in the final prompt is shown in Fig. \ref{fig:prompt_freq}. Finally, we use Sora \cite{sora}, Hunyuan \cite{hunyuan}, Pixverse, Seaweed \cite{seaweed}, Kling \cite{kling}, Hailuo \cite{hailuo}, Runway \cite{runway2024gen3}, and Wan2.1 \cite{wan2.1} to generate videos based on these prompts and conduct the benchmark.

To evaluate the alignment between agent-based evaluation and human preferences, we hire 20 annotators to score over 5,500 generated videos. For each video, the annotators need to judge whether the generated content aligns with the prompt instructions. If the generated content fully aligns with the instruction of a certain dimension, the score of this dimension is 1, on the contrary, it is 0, and if it is partially met by the instruction, the score is 0.5. For dimensions rated as 0 or 0.5, the annotators should provide a detailed explanation. Finally, we could calculate the alignment between the agent's evaluation scores and human ratings.
\section{Dynamic Agent-based Framework}
\begin{figure*}[t]
	\centering
        \scriptsize
	\begin{overpic}[width=0.95\linewidth]{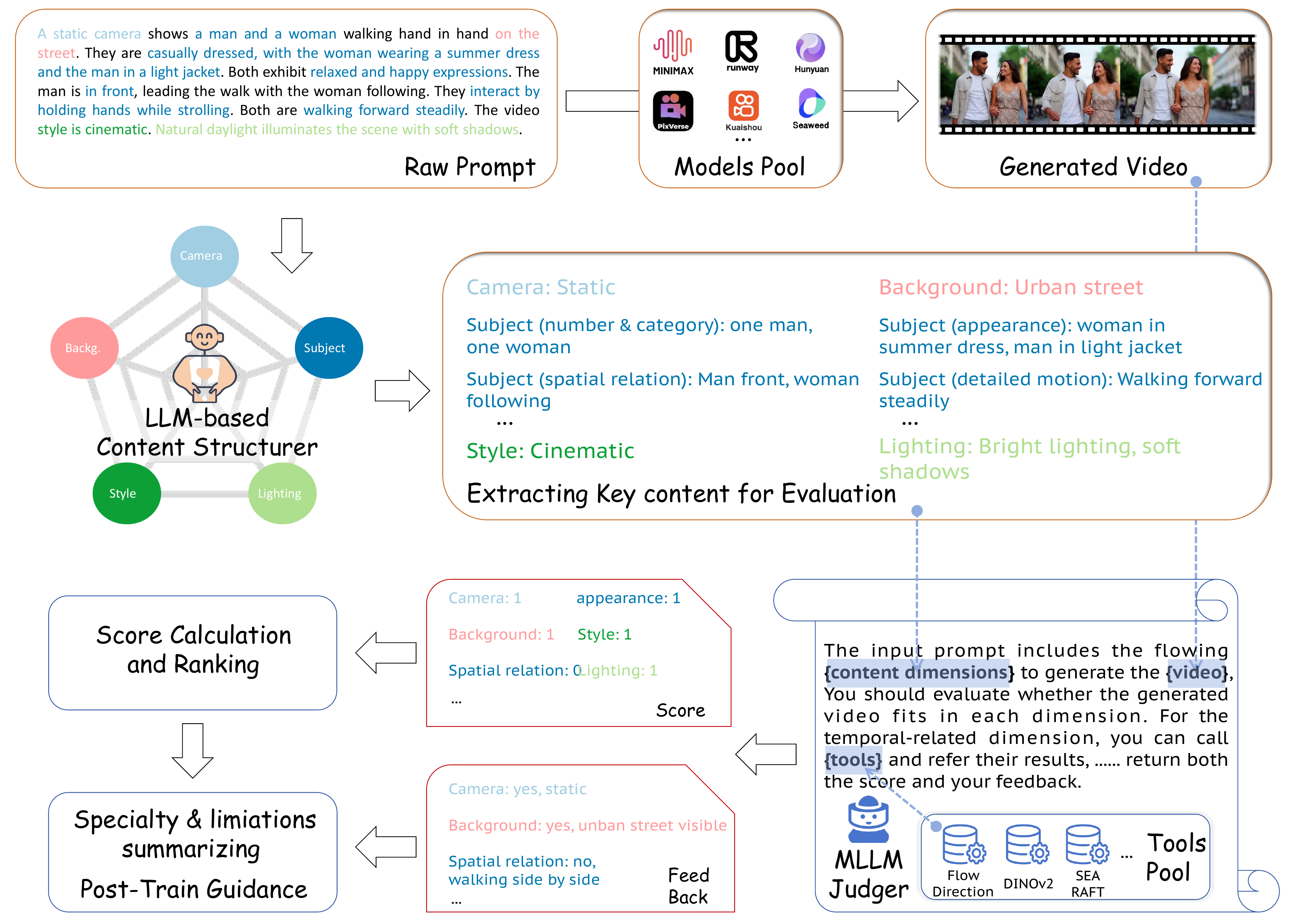}

	\end{overpic}
	\caption{\textbf{Pipeline overview.} The agent-based evaluation system is mainly composed of three parts: LLM-based content structure, MLLM-based judged, and patch tools. The content structurer parses the input prompt into dimension-specific content and sends it, along with the generated video, to the MLLM-based content judger. Leveraging the MLLM fundamental objective understanding capabilities and externally invoked temporally dense tools, the system assesses whether multiple dimensions of the input are accurately generated. The resulting scores and feedback are used for ranking, evaluation, and potentially supporting post-training.}
 \label{fig:method}
 \vspace{-0.3cm}
\end{figure*}

The overall agent pipeline is shown in Fig \ref{fig:method}. It includes prompt decomposition for content structuring and MLLM-based content judgment. Our goal is to build an agent system that can decompose input prompts into clear content blocks, then take MLLM as a foundation, and combine patch tools to achieve dynamic and more reliable evaluation of video generation.

\subsection{Content Structurer}
\label{Sec:4.1}
The prompts used for video generation are highly diverse, and directly feeding raw prompts into the MLLM for evaluation often leads to unstable and inconsistent assessment results. To further unlock the fundamental visual comprehension capabilities of the MLLM and systematically structure specific evaluation dimensions, we introduce LLMs \cite{gpt4,qwen} as the first ring of the system, leveraging their powerful text parsing capabilities to decompose the input prompts into dimension-specific contents. This transformation converts disorganized prompts into a unified, structured format, providing concise and effective assessment inputs for the MLLM, which ensures more effective and reliable outputs in subsequent stages.

\subsection{Multimodal Judger}
Employing independent operators to evaluate distinct dimensions is a common approach in existing benchmarks \cite{vebench, huang2024vbench++}, but the scalability of this method is limited. Besides, for certain objective dimensions, operators are trained on specific datasets, resulting in ambiguous scores when encountering OOD cases. Video generation is evolving at a breakneck pace, and its evaluation methodologies must advance in tandem. This necessitates an expandable and updatable evaluation framework. A viable approach involves leveraging MLLMs as the foundation to build an agent system, supplemented by specialized operators as patch tools to address gaps in the MLLMs' capabilities.

Multimodal understanding and generation models are advancing in parallel. As can be seen from the multimodal benchmark \cite{yue2023mmmu,yue2024mmmu}, commercial models already demonstrate robust performance in dimensions like semantics and appearance assessment. These models now rival or even surpass traditional operators (\eg, CLIP score \cite{clip}) in specific tasks, signaling a paradigm shift toward holistic, model-driven evaluation systems. In light of this, we develop an agent framework centered on the MLLM, we define assessments for objective dimensions as judgments rather than absolute numerical scores, to reduce the solution space and ensure more objective evaluations. The agent systematically evaluates whether the generated video fulfills each dimension's instructions by analyzing the structured prompt (Sec. \ref{Sec:4.1}) and sampled video frames. The answer includes ``yes" (1), ``no" (0), and ``half" (0.5), in which ``yes" represents a complete match with the instruction, ``no" is the opposite, and "half" indicates that part of the instruction is met. Additionally, a reason is required for each answer, aligning with the human annotation process. This not only allows us to observe the basis for the agent's scoring but may also support post-training \cite{liu2025improving}.

To improve the accuracy of the assessment, we provide detailed specifications for the evaluation criteria in each dimension and enforce a standardized and structured output format, ensuring consistent and reliable evaluations. Although it already delivers plausible results for many dimensions, challenges remain in dense-temporal evaluations since input frames for the MLLM are not temporally dense, and the collapse in certain frames is hard to capture. Therefore, the agent needs to call external tools to improve the reliability of such dimensions. We organize and further improve some operators that calculate temporal-dense metrics for agents to call. In this way, the agent can use sparse frames to judge objective dimensions, and can also call patch tools to further enhance the evaluation of temporal dense dimensions.

\begin{figure*}[t]
	\centering
        \scriptsize
	\begin{overpic}[width=0.93\linewidth]{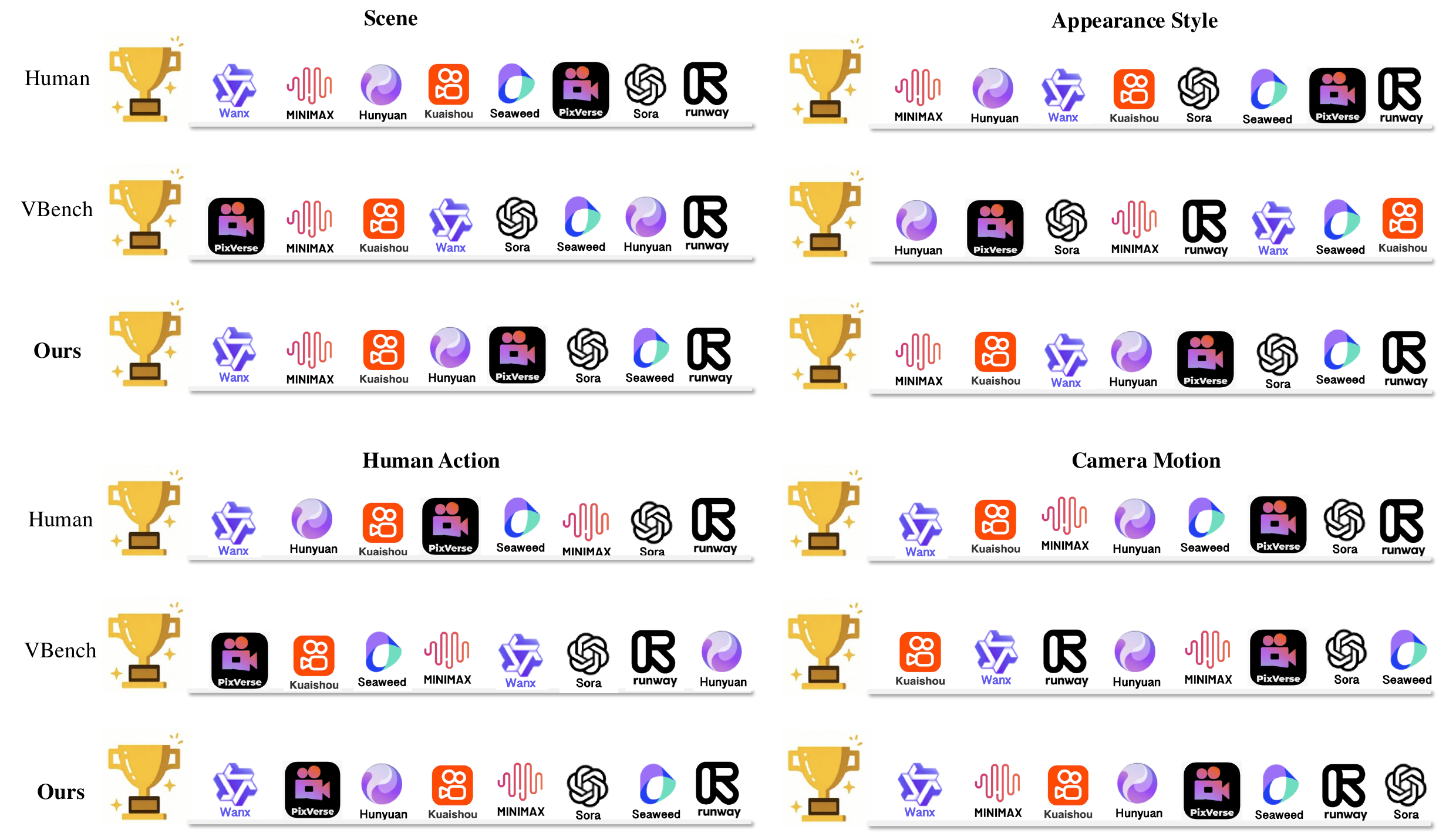}
 
	\end{overpic}
	\caption{Comparisons among Vbench operators, our agent system, and human rankings on several evaluation dimensions.}
 \label{fig:compare}
\end{figure*}

\subsection{Temporal Patch Tools}
\paragraph{Temporal Consistency.}
Assessing temporal consistency remains a critical challenge in evaluating the quality of AI-generated videos. Prior methods such as~\cite{tgve, evalcrafter,vbench} employ averaged frame-wise similarity metrics based on CLIP, DINO, etc. However, these approaches prioritize semantic consistency over reasonable motion variations, favoring static scenes while struggling to effectively evaluate plausible dynamic changes.  Alternative methods leverage low-level motion information for temporal consistency measurement, \eg, optical flow magnitude \cite{evalcrafter} and tracking results \cite{vbench, t2vcomb, chronobench}. Though these pixel-level analysis methods partially address the issue, they face inherent challenges:
1) the smoothing effect derived the direct average on spatial-temporal dimensions, which tends to mask localized distortions or abrupt video changes; 2) interference susceptibility from camera movements and scene transitions.

To alleviate these issues, we adopt a general modification instead of directly adopting the commonly used operators like CLIP and flow scores. Specifically, to tackle the smoothing effect, we first divide the video into different local units along both the temporal and spatial dimensions. Then, the variations of each spatial patch across different temporal windows are calculated based on a sliding window. For each patch, the maximum variation across all temporal windows is taken as the score for that patch. This process allows the measurement with the most variation for each patch in different temporal contexts. The average ratio of each patch exceeding the median value is the final score, which can be represented as:
\begin{align}
    s &= \{\max_{1 \le j \le W}(f(p_{j}^{i})) \mid i \in N\}, \\
    o &= \frac{1}{M} \sum_{i=1}^N \frac{s_i-\gamma}{\gamma} \mathbb{I}_{s_i>\gamma},
\end{align}
where $p_{j}^{i}$ represents the $i_{th}$ spatial patch in the $j_{th}$ sliding window, and $W$ is the number of sliding windows. $N$ is the number of spatial patches. $\gamma$ refers to the median in $s$ and $\mathbb{I}$ is the indicator function. $M$ denotes the number of patches that have greater scores than $\gamma$. $f$ is the applied operator. Notably, this method is not restricted to a specific operator. In practice, we use the flow magnitude difference and the variance of flow direction consistency to measure anomalies given its rich dense motion information. This can be formulated as:
\begin{align}
    u(x) &= \left\|\mathbf{F}_{x_{t+1}}\right\|-\left\|\mathbf{F}_{x_{t}}\right\|, \\
    v(x) &=\frac{1}{N} \sum_{t=0}^{N-1}\left(w(x_t)-\bar{w}\right)^2, \\
    f(x) &= \alpha \cdot \eta(u(x)) + \beta \cdot \eta(v(x)),
\end{align}
where $x_t$ denotes the $t_{th}$ frame and $\mathbf{F}_{x_t}$ denotes the corresponding flow. $w$ calculates the cosine similarity between adjacent frames and $\eta$ represents to the min-max normalization. $\alpha$ and $\beta$ are fixed weights, set to 0.5 in practice.

\begin{figure*}[t]
	\centering
        \scriptsize
	\begin{overpic}[width=0.92\linewidth]{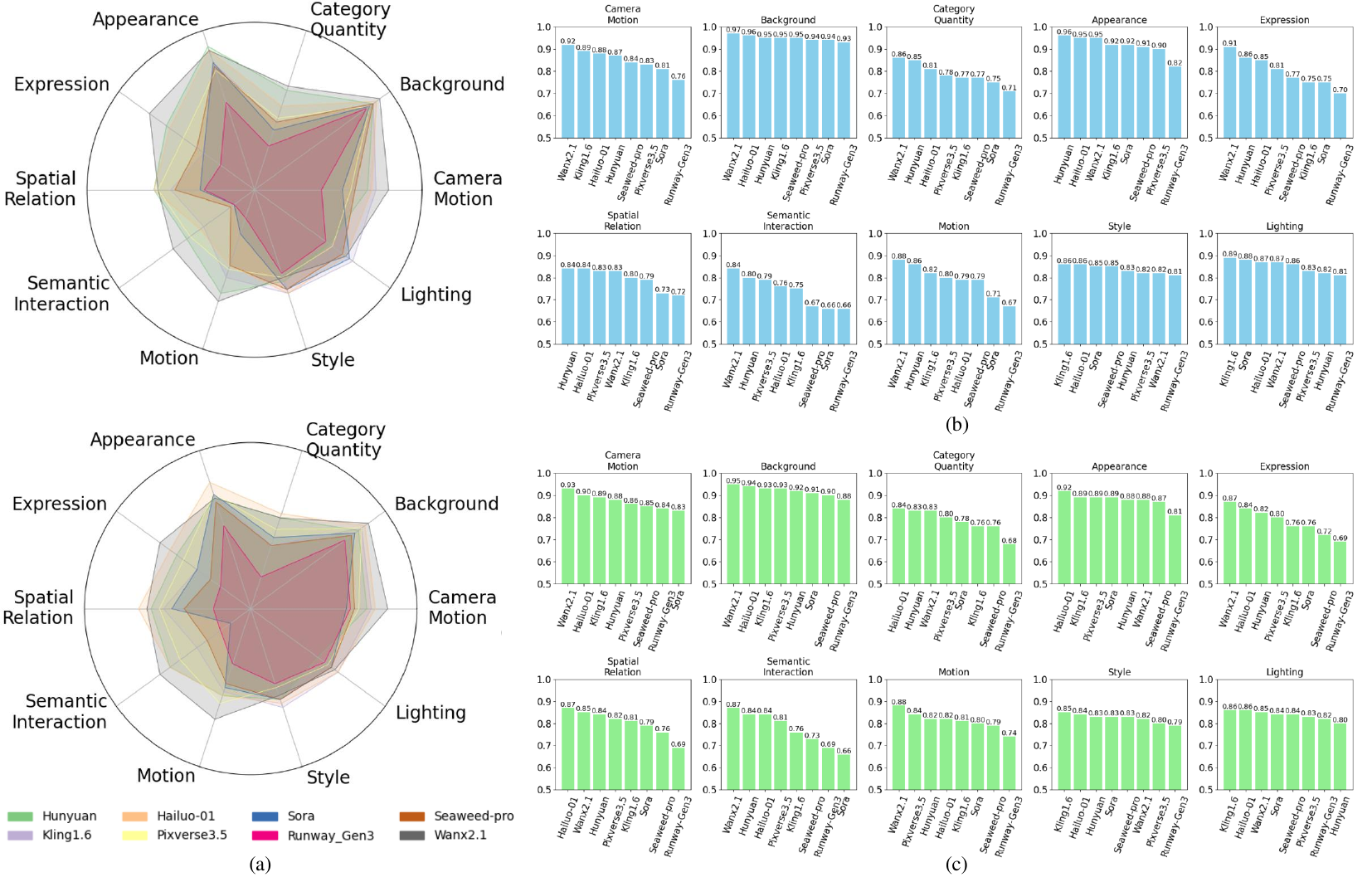}
 
	\end{overpic}
	\caption{\textbf{(a)} The distribution of scores given by humans and agent systems to each dimension of 8 models (The top denotes the human, and the bottom denotes the agent system). As well as the specific scores, \textbf{(b)} denotes the human, and \textbf{(c)} denotes the agent.}
 \label{fig:score}
\end{figure*}

\begin{figure}[t]
	\centering
        \scriptsize
	\begin{overpic}[width=1.\linewidth]{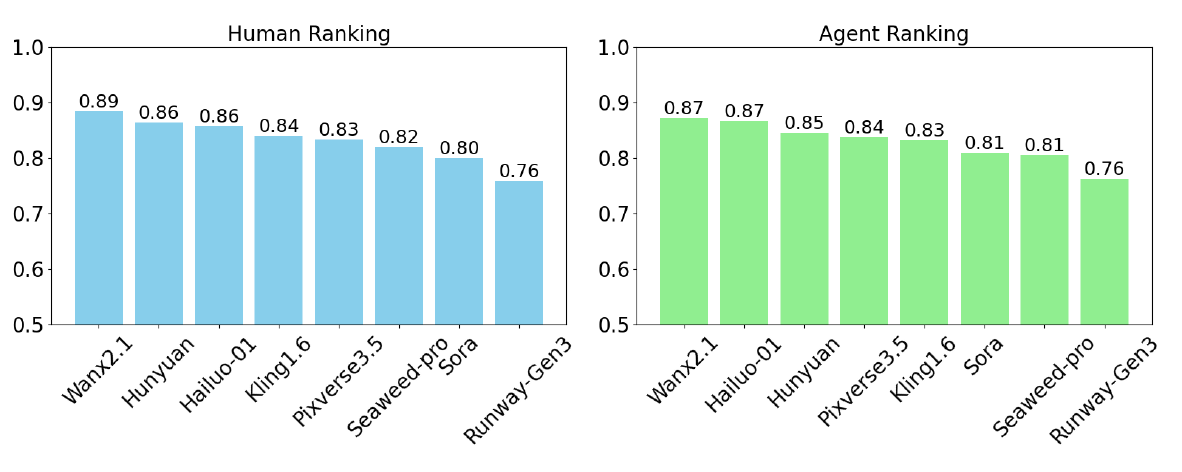}
 
	\end{overpic}
	\caption{Overall ratings and rankings of various models by humans and the agent evaluation.}
 \label{fig:align}
\end{figure}

To eliminate the effects of camera motion and scene transitions, we leverage a temporal event-splitting strategy. In specific, input frames are projected into the HSV space, and temporal differences of frames exceeding a certain threshold are marked as boundary frames. Inspired by~\cite{transition}, the HSV space is insensitive to lighting changes, thus ignoring factors like light intensity, and focusing on transitions or significant motion instead. After that, we utilize the camera motion compensation strategy. The global camera motion is estimated based on the homography transformation and subsequently removed from the overall motion.

\textbf{Subject Consistency.}
Given the excellent and consistent subject representation of DINOv2~\cite{dinov2} in a series of downstream tasks such as semantic segmentation~\cite{liu2025simple, wei2024stronger, nguyen2023cnos} and depth estimation~\cite{zeng2024wordepth, leduc2024soccernet}, we adopt it instead of DINO~\cite{dino} to assess the subjective consistency in the video.

\textbf{Dynamic Degree.}
When measuring the dynamics of the video, the accuracy of the estimated flow is essential. In this work, we replace RAFT~\cite{raft} with SEA-RAFT~\cite{searaft} due to its superior performance in occluded pixels, textureless regions, fast motion, and cross-domain generalization.

Beyond directly using operator scores as metrics, we structurally organize and provide the foundation MLLM with each operator's score, its interpretation, and the quality indications associated with their score ranges for judgment, thereby further enhancing the evaluation reliability in temporal-dense dimensions.
\section{Experiment}
\textbf{Implementation}. In the evaluation, we employ the GPT-4o \cite{gpt4} as the content structurer and qwen-vl-max \cite{Qwen-VL} as the MLLM judger to construct the agent system. For objective dimension evaluation, we uniformly extract 8 frames from each generated video. For temporal-dense dimensions, we take operators to pass all frames of the entire video to obtain scores, which are then passed to the MLLM. Finally, we compare the agent score with the human score, for each dimension of a generated video, if the two scores are consistent, it is considered aligned. If one score is $0.5$ and the other is $1$, a weight of $0.5$ will be given. Otherwise, it is considered misaligned.

\subsection{Comparison with Existing Benchmark}
We use operators from Vbench \cite{vbench} to calculate scores for defined objective dimensions, and rank the model scores for each dimension. We compare the Vbench operator and agent ranking with the human ranking in each dimension to demonstrate the superiority of the agent over previous operators. As shown in Fig. \ref{fig:compare}, the ranking of the agent scores is generally close to that of humans, \eg, Top-4 in the scene and camera dimension. However, the results calculated by Vbench operators are far from the human preferences. 

\subsection{Human Alignment Evaluation}
We show the degree of alignment between agent evaluation and human preferences through the distribution (Fig. \ref{fig:score} (a)), the average score in each dimension (Fig. \ref{fig:score} (b) and (c)), and overall score and ranking of the human and agent system across all models (Fig. \ref{fig:align}). Plus, we provide the alignment degree between the agent system and human evaluation on each dimension of each model, please refer to the appendix Fig. \ref{fig:supp_alignment}. Through these aspects, we can see that the agent system demonstrates a high degree of alignment with human preferences, indicating its reliability.

\subsection{Image-to-Video Evaluation}
We employ the same system to evaluate image-to-video generation, except that we specify the prompt image as a separate input. Unlike T2V, some objective dimensions, \eg, the appearance, background, etc. are already determined by the input image in I2V generation. Thus, we focus on the evaluation of the motion dimension, including camera motion, subject interaction, and motion details. The evaluation scores of each model through the agent system are shown in Tab. \ref{tab: i2v}.

\begin{table}[]
\footnotesize
\renewcommand{\arraystretch}{1.}
\renewcommand{\tabcolsep}{3pt}
\begin{tabular}{c|ccccccc}
\hline
               & \textbf{Wan2.1} & \textbf{Kling} & \textbf{Hailuo} & \textbf{Pixvers.} & \textbf{Seawe.} & \textbf{Sora} & \textbf{Runway}  \\ \hline
\textbf{camera.}    & \textbf{0.87}        & 0.82       & 0.84        & 0.81         & 0.84         & 0.79         & 0.76               \\
\textbf{Inter.}  & 0.78        & \textbf{0.84}        & 0.72        & 0.73         & 0.75         & 0.74         & 0.76                  \\
\textbf{Mot.} & \textbf{0.86}        &   0.84      & 0.85        & 0.81         & 0.83         & 0.84         & 0.75                  \\ \hline
\end{tabular}
\caption{The ratings of Image-to-Video Evaluation by the agent system, Mot. denotes the motion detail.}
\label{tab: i2v}
\end{table}

\begin{table}[]
\scriptsize
\renewcommand{\arraystretch}{1.}
\renewcommand{\tabcolsep}{4pt}
\begin{tabular}{c|cccccccccc}
\hline
               & \textbf{Ca.} & \textbf{Bg.} & \textbf{CQ.} & \textbf{Ap.} & \textbf{Ex.} & \textbf{Sp.} & \textbf{In.} & \textbf{Mo.} & \textbf{St.} & \textbf{Li.} \\ \hline
\textbf{\ding{55}}    & 0.64        & 0.75        & 0.54        & 0.56         & 0.32         & 0.40         & 0.46         & 0.51         & 0.50         & 0.56         \\
\textbf{Qwen}  & 0.78        & 0.87        & 0.78        & 0.65         & 0.60         & 0.63         & 0.70         & 0.74         & 0.74         & 0.73         \\
\textbf{GPT-4o} & 0.75        & 0.90        & 0.74        & 0.62         & 0.63         & 0.62         & 0.71         & 0.69         & 0.72         & 0.70         \\ \hline
\end{tabular}
\caption{Human alignment ratio with (Qwen, GPT-4o) or without (\ding{55}) structured content. Ca. denotes the camera motion, Bg. is the background, CQ. is the category and quantity, Ap. is the appearance, Ex. is the expression, Sp. is the spatial relation, In. is the interaction, Mo. is the motion detail, St. is the style and Li. indicates the lighting.}
\label{tab: content}
\end{table}

\subsection{Ablation Study}
\textbf{Structured Content}. Decomposing the original input prompts into structured content and delivering clear evaluation criteria to the MLLM is pivotal to the whole system. We conduct experiments to verify the effectiveness of structuring contents from raw prompts into dimension-specific content through the LLM \cite{gpt4,qwen}. As shown in Tab. \ref{tab: content}, directly feeding raw prompts into the MLLM may causes conceptual confusion, resulting in weaker alignment with human preferences. 

\noindent\textbf{Temporal Patch Tools}. Due to the limited frames that are fed to the MLLM, the model still has a gap in temporal-dense perception compared to humans. We employ patch operators as tools to compensate for this to a certain degree. Shown in Tab. \ref{table:tool} (a), the integration of temporal tools notably improved the alignment with human preferences in temporal-dense dimensions. Furthermore, our improvements to the operator further enhance the alignment ratio. Additionally, we test to use semantic operators (\eg, CLIP score \cite{clip}) as tools for specific dimensions (shown in Tab. \ref{table:tool} (b)). The results indicate that these operators reduce alignment degree, further demonstrating their unreliability for video generation evaluation.

\begin{table}[t]
% \scriptsize
\footnotesize
  \renewcommand{\arraystretch}{1.}
  \renewcommand{\tabcolsep}{5.pt}
\label{table:ablation}
\begin{subtable}[t]{0.4\linewidth}
\begin{tabular}{c|ccc}
\toprule
\textbf{} & \textbf{Ca.}  & \textbf{In.} & \textbf{Mo.}  \\ \midrule
\textbf{\ding{55} T.}    & 0.61     & 0.51     & 0.54     \\
\textbf{Base T.}    & 0.71     & 0.62    & 0.68     \\
\textbf{Impr. T.}    & 0.78     & 0.70    & 0.74     \\
\bottomrule

\end{tabular}
\caption{}
\end{subtable}
\begin{subtable}[t]{0.7\linewidth}
\centering
\begin{tabular}{c|ccc}
\toprule
\textbf{} & \textbf{Bg.}  & \textbf{CQ.}  & \textbf{Ap.} \\ \midrule
CLIP    & 0.82     & 0.72     & 0.58     \\
Ours   & 0.87     & 0.78     & 0.65   \\
\bottomrule
\end{tabular}
\caption{}
\end{subtable}
\caption{{\textbf{(a)} Human alignment ratio in temporal-dense dimension without (\ding{55}) tools (T.), using base tools and our improved tools. \textbf{(b)} Human alignment ratio of objective dimension when introducing CLIP as the tool.}}
\label{table:tool}
\end{table}

\begin{figure}[t]
	\centering
        \scriptsize
	\begin{overpic}[width=1.\linewidth]{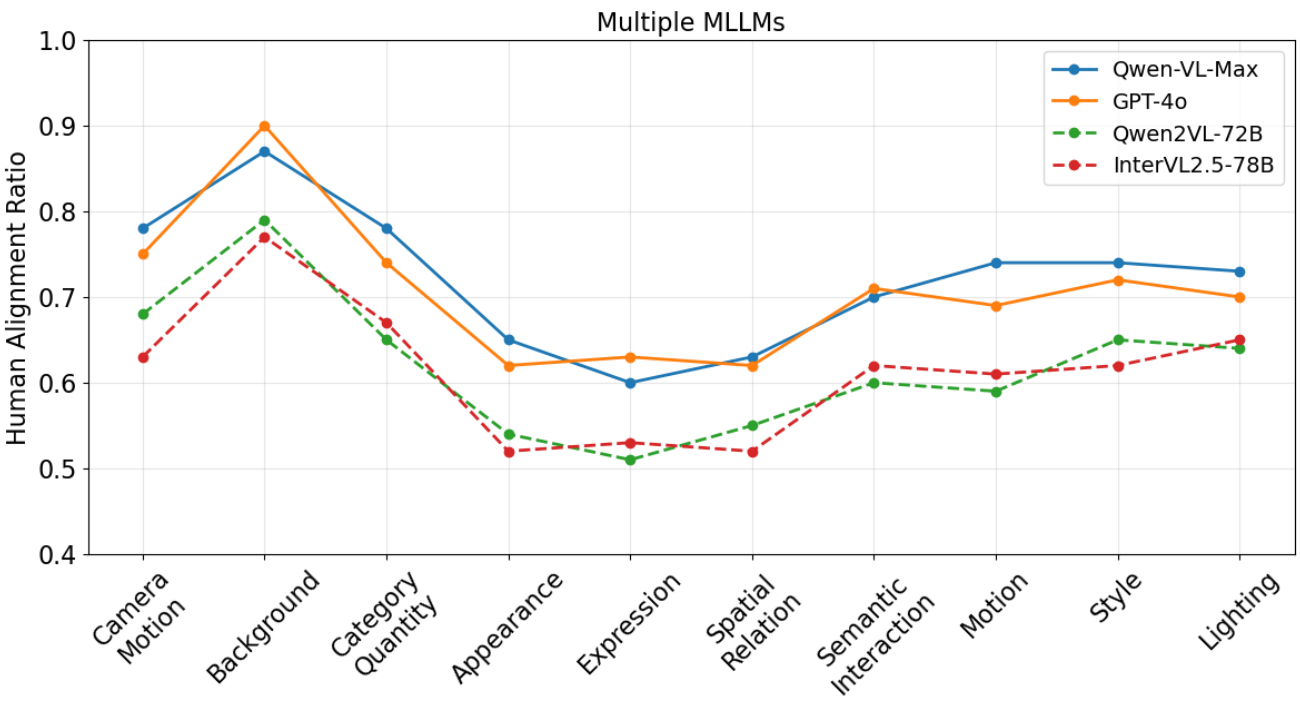}
 
	\end{overpic}
	\caption{Human alignment ratios with different base MLLMs.}
 \label{fig:agentbase}
\end{figure}

\noindent\textbf{Agent Base}. Our agent-based evaluation system is dynamic, the entire evaluation framework is designed to be flexible and extensible. We test the performance of various base MLLMs, like QwenVL-72B \cite{Qwen-VL}, InterVL-78B \cite{chen2024internvl} and closed-source (GPT-4o \cite{gpt4}, with results shown in Fig. \ref{fig:agentbase}. From the gap between the open-source and closed-source models, it can be seen that MLLMs continue to evolve, and the reliability of this agent-based evaluation system could also be further enhanced.
\section{Conclusion}
We introduce VideoGen-Eval, a novel agent-based dynamic evaluation framework that integrates structured content, MLLM-based judgment, and patch tools to assist in addressing the gap between recent video generation and evaluation. Coupled with our benchmark, this evaluation system exhibits a comprehensive and human-aligned evaluation of cutting-edge video generation models. This establishes a dynamically extensible paradigm for video generation evaluation. As MLLMs and video generation models continue to co-evolve, the proposed system framework enables more robust evaluation capabilities while progressively aligning with human preferences.

\begin{figure*}[!h]
	\centering
        \scriptsize
	\begin{overpic}[width=1.\linewidth]{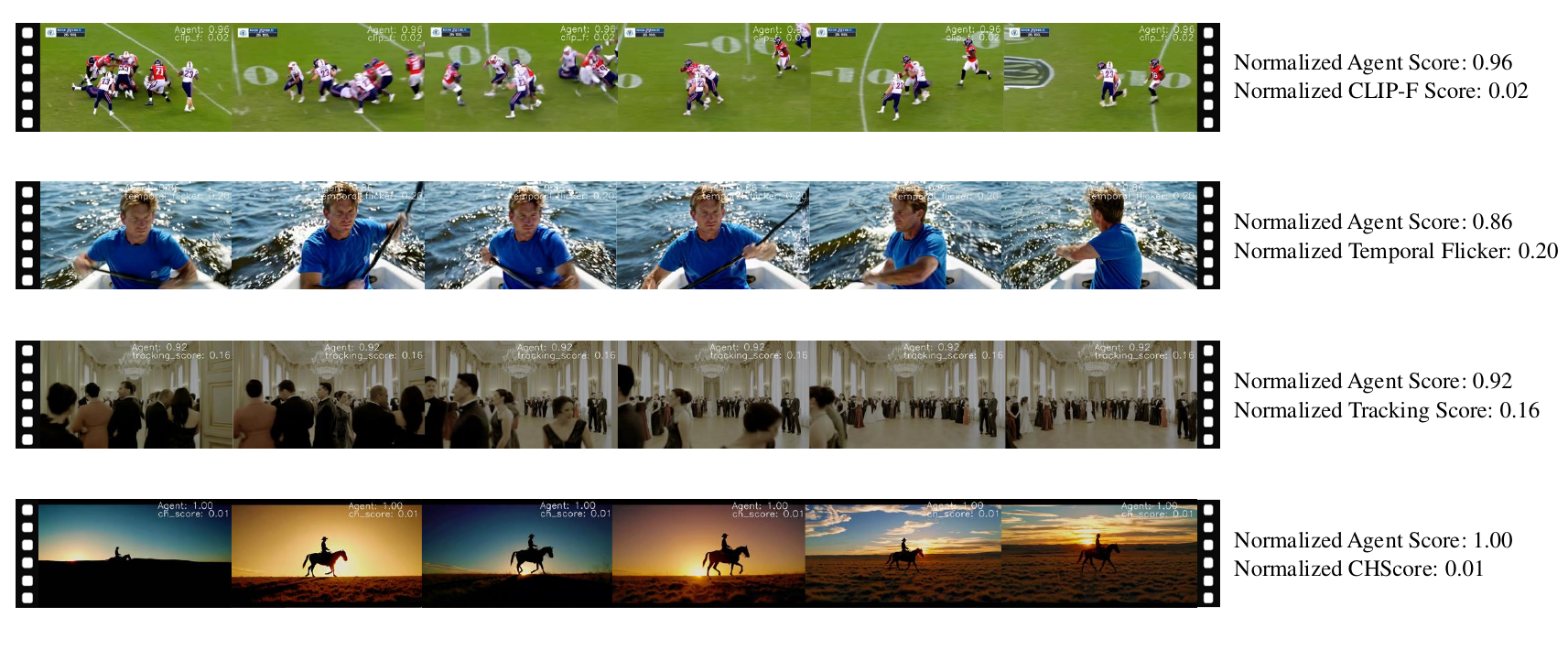}
	\end{overpic}
	\caption{Comparison of scores under different measurements.}
 \label{fig:score_compare}
\end{figure*}

\noindent\textbf{Limitations.} Although currently agent system could evaluate several major dimensions of generated video content, it is still difficult to achieve expert-level judgments in specialized domains. This may be due to the lack of fine-grained annotations during the training of base MLLMs. A promising future direction involves creating high-quality annotated datasets and fine-tuning expert MLLMs for evaluating the generated content, or developing specialized models trained on domain-specific small-scale data as patch tools, which then could be utilized within the system to enable higher expert-level evaluation.
\begin{figure*}[!ht]
	\centering
        \scriptsize
	\begin{overpic}[width=0.97\linewidth]{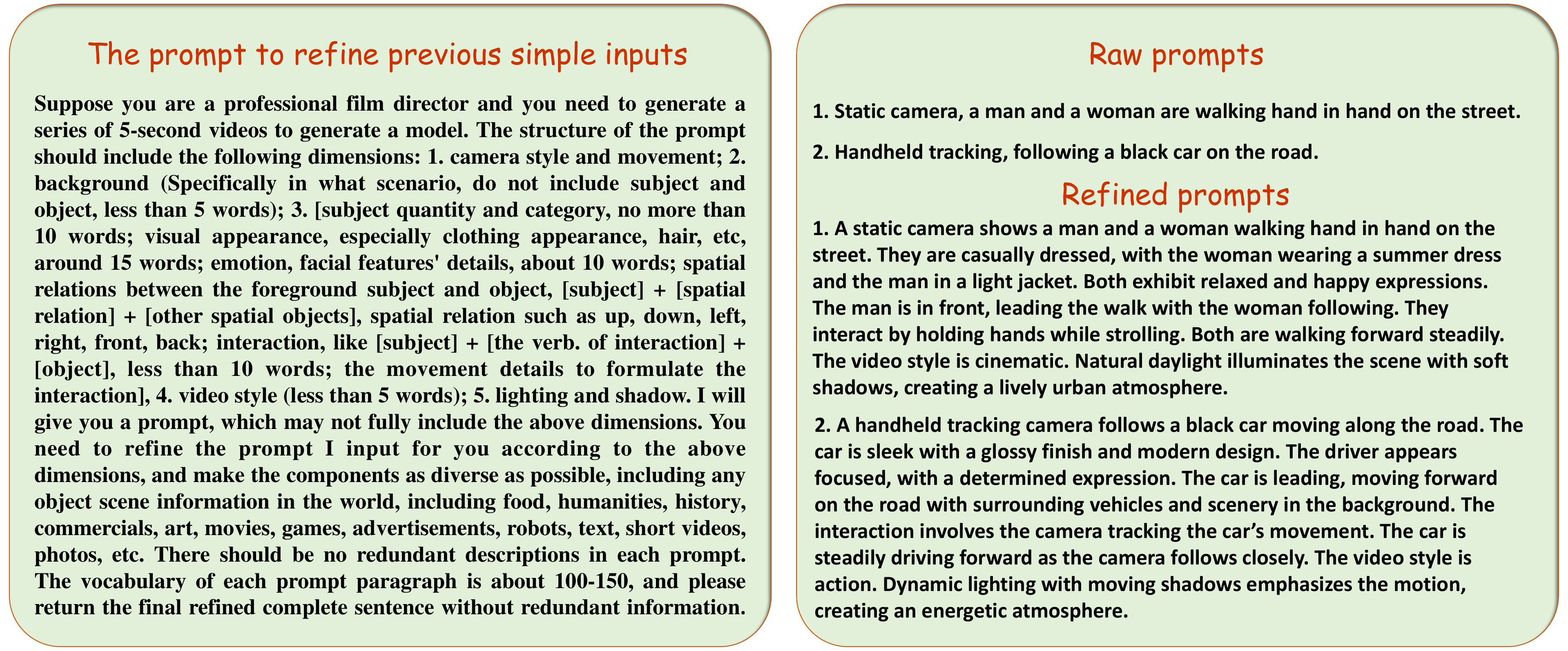}
	\end{overpic}
	\caption{\textbf{Left.} The instruction we use for GPT-4o to refine some simple prompts in existing benchmarks; \textbf{Right.} Comparison of the raw prompts and refined prompts.}
 \label{fig:supp_extend_prompt}
\end{figure*}

\begin{figure*}[!ht]
	\centering
        \scriptsize
	\begin{overpic}[width=0.97\linewidth]{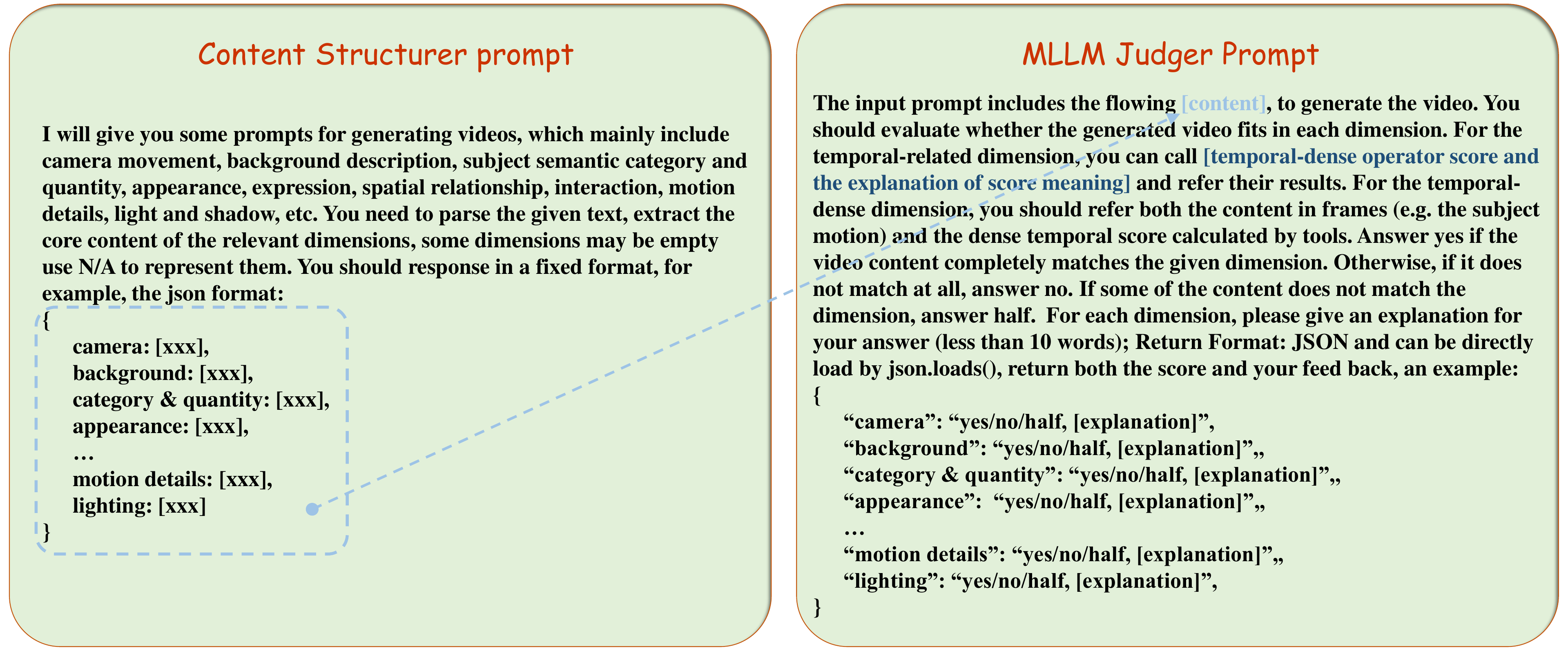}
	\end{overpic}
	\caption{\textbf{Left.} The prompt and output format for the content structurer; \textbf{Right.} The prompt and output format for the MLLM judger}
 \label{fig:supp_prompt2}
\end{figure*}
\section{Appendix}
\appendix
\section{Comparison with Previous Measurements}
\begin{figure*}[!h]
	\centering
        \scriptsize
	\begin{overpic}[width=1.\linewidth]{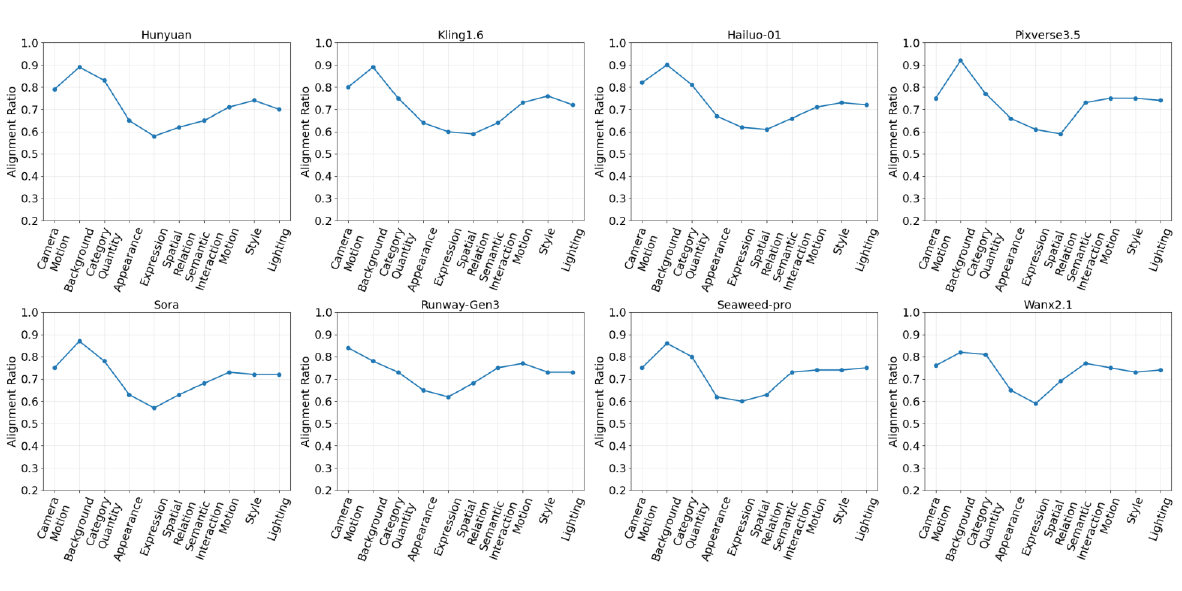}
	\end{overpic}
	\caption{Alignment ratios of agent evaluation to human evaluation on different models across multiple dimensions.}
 \label{fig:supp_alignment}
\end{figure*}

In this section, we further compare the proposed evaluation Agent with partial strategies adopted in previous benchmarks. We reference the CLIP-F operator commonly used in prior works to measure inter-frame consistency, the tracking score~\cite{cotracker} based on tracking strategies, CHScore~\cite{chronobench}, and the temporal flicker operator in VBench~\cite{vbench} for measuring temporal flickering. Among these, the tracking score directly measures motion quality using the proportion of non-lost tracking points. It first calculates the subject mask using U2Net~\cite{u2net}, then computes the failure rate of pixel tracking within the subject area. CHScore~\cite{chronobench} measures the temporal coherence based on the standard deviation and extreme value changes of the ratio of lost points over time. Specific comparison details are illustrated in Figure~\ref{fig:score_compare}. More details are available in the supplementary materials. During comparison, considering different operators have varying scales, we applied min-max normalization to rescale all scores to [0,1] to eliminate scale differences. Therefore, some scores may approach zero, reflecting relative rankings rather than absolute video quality.

Through observations, we identify that these operators may function in specific scenarios but face challenges: CLIP-based temporal consistency metrics tend to favor static scenes while struggling to effectively evaluate reasonable high-quality vigorous motions, large movements, or camera motions, often assigning lower scores. The tracking score (Note: All displayed scores are min-max normalized rather than absolute values) for subject pixel motion is susceptible to camera movement interference. Although segmentation methods partially mitigate background/camera motion effects, current tracking models remain vulnerable to significant camera changes or scene transitions. VBench's temporal flicker operator proves reliable for simple low-quality motions but struggles with scenarios involving frequent local pixel variations (e.g., water splashes in boat-rowing videos). It also tends to misjudge camera movements and intense motions, resulting in abnormally low scores. In contrast, our Agent demonstrates more reliable results through human perception alignment and comprehensive multi-dimensional evaluation.

\section{Craft Prompts}
We merge a part of the original prompts from existing benchmarks as the final test sets, however, some of them are originally quite simple and do not meet our requirements for prompts, so we design the instruction for GPT-4o to expand these prompts, the specific instruction is shown in the Fig. \ref{fig:supp_extend_prompt}. We also provide some examples of both the original and expanded prompts, still shown in Fig. \ref{fig:supp_extend_prompt}. It can be seen that the expanded prompts have more content, which is suitable for testing the capabilities of cutting-edge video generation models.

For the structuring of the content and the prompts fed into the foundation MLLM, we also meticulously craft the prompts and strictly standardize the output format. This ensures the stable operation of the entire agent system without the need for excessive handling of output format issues. The prompt for content structuring is shown on the left of Fig \ref{fig:supp_prompt2}, and the prompt for the MLLM is shown on the right of Fig \ref{fig:supp_prompt2}. Please note that structured content is also fed into the MLLM, and for the patch operator, we give the introduction of its function as well as the specific numerical range of its score and what it represents, all of which serve as references for the MLLM in making final judgments.

\begin{figure*}[!ht]
	\centering
        \scriptsize
	\begin{overpic}[width=1.\linewidth]{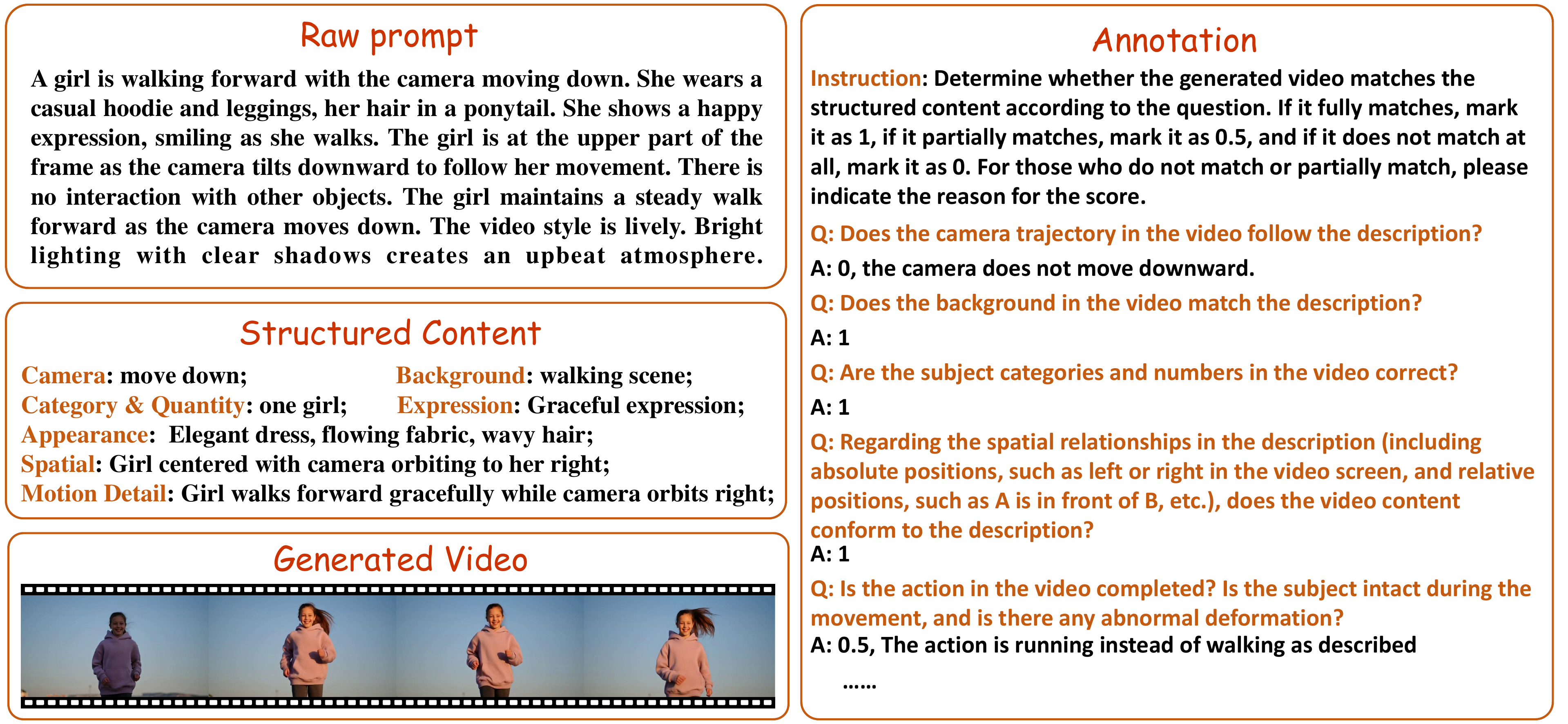}
	\end{overpic}
	\caption{Information provided in the human annotation process, as well as the annotation instruction and result examples.}
 \label{fig:supp_annoatation}
\end{figure*}

\section{Alignment between Agent and Human Evaluations}
This section further compares the alignment between the proposed agent system and human evaluation, as shown in Fig~\ref{fig:supp_alignment}, which presents alignment ratios between agent and human evaluation for different video generation models across multiple dimensions. 

Most dimensions achieve alignment ratios above 0.6. Notably stronger alignment is observed in Camera Motion, Background, Category and Quantity, etc. This demonstrates the agent's effectiveness in capturing human preferences across critical visual aspects. At the same time, we can see that there are still differences in how agents evaluate different dimensions, and these differences are consistent across different models, which also indirectly reflects the current shortcomings of MLLMs in certain dimensions. We partially address this problem through the injection of patch tools, but it is promising that as MLLMs evolve, they will gradually align with human preferences in these dimensions. Our dynamic and evolving agent-based evaluation system is expected to become more reliable.

\begin{figure*}[!ht]
	\centering
        \scriptsize
	% \begin{overpic}[width=1.\linewidth]{Fig/supp_case.pdf}
	\begin{overpic}[width=1.\linewidth]{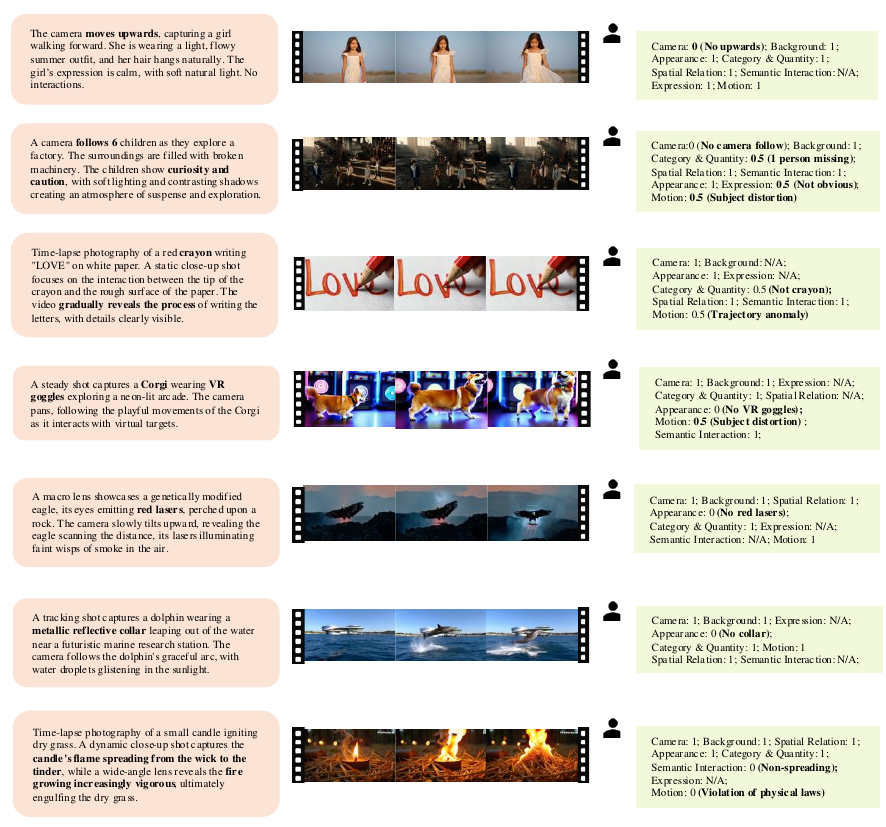}
	\end{overpic}
	\caption{Cases of the generated video and corresponding human feedback.}
 \label{fig:supp_case}
\end{figure*}

\section{Human Annotation}
For human annotation, the rules we specify are objective judgments rather than subjective conjectures within a score range, which can eliminate annotators' subjective biases and better verify the effectiveness of the agent-based evaluation system. The specific annotation process is shown in Fig \ref{fig:supp_annoatation}. We provide annotators with refined prompts and structured content. The first step is to verify whether the content structure extraction of the original prompt using the LLM is correct, and if not, correct it. After that, we will compare the generated video with the structured content for annotation. For each dimension, we provide a question and ask annotators to annotate according to the question. For dimensions that have not been done in the generated video, an explanation is required to make it clearer why the generated video does not meet the requirements.

We present examples of generated videos along with corresponding human scores and related feedback, as shown in Fig~\ref{fig:supp_case}. For each specific dimension, the maximum score is 1 point, and a score of 0 points indicates that the requirement is not fully met. For cases that do not fully meet the requirements or receive a score of 0, participants are asked to provide corresponding reasons when scoring. As illustrated in Fig~\ref{fig:supp_case}, issues such as camera movement, facial deformation, or the incorrect number of generated items lead to a decrease in the score for the corresponding dimension.

\clearpage
{
    \small
    \bibliographystyle{ieeenat_fullname}
    \bibliography{main}
}

\end{document}